\documentclass{article}

\usepackage{algpseudocode}
\usepackage{algorithm}
\usepackage{graphicx}
\usepackage[english]{babel}
\usepackage{authblk}
\usepackage{amsmath}
\usepackage{subcaption}
\usepackage{adjustbox}
\usepackage[colorlinks,allcolors=blue]{hyperref}

\usepackage{lineno,hyperref}

\title{Optimizing Dense Feed-Forward Neural Networks\thanks{Accepted manuscript in Neural Networks. DOI:10.1016/j.neunet.2023.12.015}}

\author[1]{Luis Balderas
\thanks{Corresponding author. luisbalru@decsai.ugr.es}}

\author[2]{Miguel Lastra}

\author[1]{Jose M. Benítez}

\affil[1]{Department of Computer Science and Artificial Intelligence, DiCITS, iMUDS, DaSCI, E.T.S.I.I.T. University of Granada, Spain}
\affil[2]{Department of Software Engineering, DiCITS, iMUDS, DaSCI, E.T.S.I.I.T. University of Granada, Spain}

\begin{document}

\maketitle

\begin{abstract}
	Deep learning models have been widely used during the last decade due to their outstanding learning and abstraction capacities. However, one of the main challenges any scientist has to face using deep learning models is to establish the network's architecture. Due to this difficulty, data scientists usually build over complex models and, as a result, most of them result computationally intensive and impose a large memory footprint, generating huge costs, contributing to climate change and hindering their use in computational-limited devices. In this paper, we propose a novel feed-forward neural network constructing method based on pruning and transfer learning. Its performance has been thoroughly assessed in classification and regression problems. Without any accuracy loss, our approach can compress the number of parameters by more than 70\%. Even further, choosing the pruning parameter carefully, most of the refined models outperform original ones. We also evaluate the transfer learning level comparing the refined model and the original one training from scratch a neural network with the same hyper parameters as the optimized model. The results obtained show that our constructing method not only helps in the design of more efficient models but also more effective ones.
\end{abstract}

%\begin{keyword}
%	neural network pruning, dense feed-forward neural network optimization
%\end{keyword}

\section{Introduction}

Over the last decade, deep neural networks have become the state-of-art technique on several challenging tasks such as computer vision \cite{comp-vis}, speech recognition \cite{sp-rec} or natural language processing \cite{nlp}. Due to the huge amount of data and hardware development and innovation in the field, larger and deeper learning models have been employed to face more complex problems and learn more elaborated patterns from data. Concretely, feed-forward neural networks, which represented the beginning of the era of neural networks, are still an extraordinary and widely used tool in solving machine learning problems with tabular data not only in the industry but also in academia. More than 58000 articles and more than 51000 patents can be found in Scopus related with this type of neural networks.

One of the main challenges any scientist has to face when using deep learning models is to establish the network's topology and its hyper parameters. Assuming that we have millions of parameters, it can be intricate to design a suitable structure for profitable learning, as we would need to choose the number of layers, the number of units and their distribution. Generally, machine learning engineers come up with enormous neural networks to assure their models are complex enough to find the correct patterns in the learning tasks they are facing. Many problems can be addressed by this methodology and modern hardware developments facilitate it (it is known that when training deep learning neural networks, graphical processing units, GPUs, can be more than two orders of magnitude faster than CPUs). However, this solution might not be the most efficient one, due to the fact that these high capacity networks have significant inference and energetic costs \cite{han2015}. In 2019, researchers at the University of Massachusetts found that training several large Artificial Intelligence models (including neural architecture search) can emit more than 284019 kilograms of carbon dioxide, in other words, nearly five times the lifetime emissions of the average American car (including the manufacture of the car itself) \cite{str2019}. 

There are some ways of designing efficient deep neural networks and generating effective solutions for a given learning task. For example, using memetic algorithms to find a good architecture to fit the task or, provided an architecture, using alternative optimization algorithms to fine-tune the connection weights. However, pruning is one of the most used methods to reduce neural networks complexity. In fact, pruning techniques have been extensively studied for model compression since 1990, when Optimal Brain Damage (OBD) \cite{OBD} and Optimal Brain Surgery (OBS) \cite{OBS} where designed. Along the past years, many other approaches have been presented in order to generate more efficient and effective neural networks in all their representations (dense, convolutional or recurrent).

Motivated by that general interest, we have developed a new feed-forward neural network constructing method called \textit{Optimizing Dense Feed-Forward Neural Network Algorithm} (\textbf{ODF2NNA}), based on pruning and transfer learning which requires minimal tuning. For each supervised learning problem, either classification or regression, we build a model to address the learning task but with a very general topology (the same number of neurons for each layer). Then, inspired by \cite{ARTICULOMALIGNO}, we define our pruning algorithm. This algorithm is designed to find an optimal subnet embedded in the original one, reducing the number of parameters and maintaining, or even improving, the accuracy results. The presented technique has only one parameter $\epsilon$, which represents the pruning level depending on the variability of each unit in a layer. A refining phase is carried out afterwards. Our experiments show that this method is effective: it produces neural networks with improved learning and generalization capabilities through the use of knowledge transfer from large networks to the pruned and refined ones.

The rest of this paper is structured as follows: In Section 2, we introduce the state-of-art of different approaches for designing efficient deep neural networks. In Section 3, we describe our proposal. In Section 4 our methodology is experimentally analyzed In Section 5 we discuss the results and Section 6 highlights the conclusions. 

\section{Previous work}

Designing a well-generalized architecture for deep neural networks is an important task.  One approach that must be considered is neuroevolution. Neuroevolution enables important capabilities such as including learning neural network building blocks, hyperparameters, architectures and even the algorithms for learning themselves. Besides, it enables extreme exploration and massive parallelization due to the fact that neuroevolution maintains a population of solutions during the search (\cite{Stanley2019}). In \cite{neat2002} NEAT algorithm is described as a classical neuroevolution approach. More recently, \cite{neat2018} took inspiration from NEAT and evolved deep neural networks by starting small and adding complexity through mutations. In consequence, entire layers of neurons are added achieving impressive performance on the CIFAR dataset. In \cite{Real_Aggarwal_Huang_Le_2019} we find a variant approach which improved performance by evolving small neural network modules that are repeatedly used in larger hand-coded blueprint, which consists in stacking the same layer module to make a DNN, like Inception, DenseNet and ResNet architectures (\cite{Stanley2019}). Another approach to be considered is Neural Architecture Search (NAS), which aims to automate the architecture designs of Deep Neural Networks. Mathematically, NAS can be modeled by an optimization problem. Based on the optimizer used, the existing NAS algorithms can be classified into three categories: reinforcement learning based NAS algorithms, gradient-based NAS algorithms and evolutionary NAS algorithms (ENAS). ENAS algorithms are particularly effective and they might be used in many different real-world applications, such as image classification, speech recognition, language modeling or traffic flow forecasting between others. In \cite{enas2021} we find a complete ENAS algorithms survey. In order to enhance the hyper-parameter selection process in Deep Neural Networks, biologically-inspired approaches should be considered. In particular, \cite{pso2017} deploys a Particle Swarm Optimization method as a wrapper to the training process to retrieve hyper-parameters that minimize the classification error. In fact, they present an extensive experimental study involving convolutional neural networks and LeNet-4 network on the MNIST dataset and some very small DNN models optimized using PSO for the CIFAR-10 dataset. Hardware optimizations for accelerating Deep Neural Networks should be considered too (\cite{hardwareopt})

Although the aforementioned methods show very promising results, they demand heavily computational resources (\cite{nap}, \cite{sepnas}). Pruning approaches are generally more efficient for deep neural networks and, as a result, they are extensively used for convolutional neural networks and feed-forward neural network optimization. More details about CNN pruning and quantization methods can be found in \cite{LIANG2021370}. Nonetheless, as ODF2NNA is designed only for feed-forward networks, CNN optimization methods are not comparable with ODF2NNA, so they are out of the scope of this paper. We will focus on feed-forward neural network optimization methods.  Sensitivity-based methods (\cite{PBM}) try to understand how important each weight or node is in the network to prune those which have the least effect on the objective function. The most important pruning algorithms of this kind are OBD and OBS. The basic idea of OBD (\cite{OBD}) is to use second-derivative information to find a trade-off between network complexity and training set error. The OBD procedure can be carried out as follows: First, choose a network architecture and train it until an accurate solution is obtained. Then, compute the second derivatives and the salience for each parameter. Finally, sort the parameters by salience, delete the lowest ones and start training again. On the other hand, OBS (\cite{OBS}) trains a neural network minimizing the error. After that, computes the inverse Hessian matrix, in order to find the parameter which gives the smallest salience. If this increment in the candidate error is much smaller than the Tailor's error function (E), then this weight should be deleted and the remaining weights are updated. Otherwise, we may have found that no more weights can be deleted without a large increase in E, so it may be desirable to retrain the network. Consequently, OBD permits pruning more weights than other methods and thus yields better generalization. Dong et al. \cite{dong2017learning} borrow ideas from OBS and OBD presenting L-OBS (Layer-wise OBS), whose intention is restrict the computation on second order derivative, i.e., the Hessian matrix is only computed for the parameters of an specific layer, making the computations tractable (resp. OBD, where the inverse of the Hessian matrix is calculated over all the parameters). Besides, they reduce computational complexity of the inverse operation of the Hessian matrix by using characteristics of back-propagation for dense layers in deep networks. 

LWC \cite{han2015} begins learning which connections are important via regular network training. Then, they prune the low-weight connections, in other words, connections with weights below a threshold are removed from the network. Finally, the network is retrained to learn the final weights for the remaining connections. Guo et al. \cite{guo2016dynamic} propose to sever redundant connections by what they call dynamic network surgery (DNS). This method involves pruning and splicing. The pruning operation is made to compress the network but over pruning or incorrect pruning might be responsible for accuracy loss. In order to compensate it, they incorporate the splicing operation so as to recover connections once the pruned connections are found to be important any time. 

Engelbrecht \cite{VNP} proposed a modified approach to perform sensitivity analysis. Instead of directly using the sensitivity value, Engelbrecht uses the average sensitivity of the network parameters which is computed over all the patterns and then a new measure called variance nullity is applied. As a result, the Variance Nullity Pruning (VNP) method allows pruning both nodes and weights. Hagiwara \cite{HAGIWARA1994207} presented a Magnitude Based Pruning (MBP) method suggesting three strategies called Goodness factor, Consuming energy and Weights power for detecting redundant hidden neurons. Augasta et al. \cite{Augasta2011ANP} proposed the Neural Network Pruning by Significance (N2PS) pruning method. N2PS is based on a measure which is calculated by the sigmoidal activation value of the node and all the weights of its outgoing connections. Every node whose significance value is below a threshold is considered insignificant and therefore eliminated. Xing et al. \cite{Xing2009} proposed a two-phase construction approach for pruning input and hidden units of dense neural networks based on mutual information. 

Han et al. \cite{han2016deep} presented a three-stage pipeline to reduce the storage required by neural networks. First, they prune the network by removing redundant connections. Next, the weights are quantized so that multiple connections share the same weights. Finally, they apply Huffman coding to take advantage of the biased distribution of effective weights. Manessi et al. \cite{8546129} introduced the following compressing method: given a neural network $N$, they build a sibling network $N_s$ that is explicitly able to shrink the trainable weights of $N$. Then, they train $N_s$ by means of a gradient descent-based technique introducing a regularization term; and finally, they build $N_p$ from $N_s$, (pruned version of the original network $N$). Guo et al. \cite{7939970} proposed a learning automata-based method to train deep neural networks prepared to prune the weakly connected units. For all models, they allocate 1000 units for each layer, and use the ReLU activation function and Gaussian initialization. 

Despite the great number of existing research papers focused on Neural Networks pruning, the state of the literature is such that it is not possible yet to answer crucial questions such as 'which techniques achieve the best accuracy/efficiency trade off?' or 'are there strategies that work best on specific architectures or datasets?' The reason is that we are suffering a lack of standardized benchmarks and metrics \cite{2020state}.

\section{Proposal}

In this paper, we address the challenge of establishing the network's topology. Thus, we introduce Optimizing Dense Feed-Forward Neural Network Algorithm (\textbf{ODF2NNA}), a new feed-forward neural network construction method based on pruning and transfer learning. In this section we pay special attention to the process that performs the identification and extraction of relevant units (called \textit{useful units}) and finally the construction of the refined model.

\subsection{ODF2NNA}

\begin{figure}[H]
	\centering
	\includegraphics[scale=0.35]{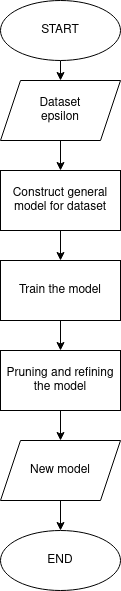}
	\caption{Algorithm 1. Flowchart}
	\label{alg1}
\end{figure}

ODF2NNA is the neural network construction method we propose. It can be used for classification and regression problems and employs a three-step process:  construct a general model, train it and build a new one pruning and refining the general model (Algorithm 1, Figure \ref{alg1}). It receives two arguments: a dataset, which feeds a supervised learning task to address, in our case, building feed-forward neural networks, and a pruning tolerance parameter, $\epsilon > 0$.   \\

\begin{algorithm}[H]
	\caption{ODF2NNA}
	\begin{algorithmic}
		\Function{ODF2NNA}{dataset, $\epsilon$}
		\State Construct a general model for addressing the \textit{dataset} learning task.
		\State Train the general model 
		\State Build a new model by pruning and refining the general model using \textit{$\epsilon$}.
		\EndFunction
	\end{algorithmic}
\end{algorithm}

Our procedure begins by receiving a dataset, which represents a classification or regression problem. Once we know the number of examples available in the dataset, we establish the number of parameters as the number of examples and build a general model composed by layers and the same number of neurons per layer. There are no restrictions imposed for the number of layers. The number of parameters of the general model depends on the number of layers chosen. Each problem requires a different number of layers to acquire an accurate model. We have used between three and ten of them.

The number of neurons per layer is determined as follows: For any hidden layer, the number of parameters is equal to the sum of the connections between layers and the bias values in every layer. In consequence, if we define

\begin{itemize}
	\item NL: number of layers.
	\item NP: number of parameters. (\ref{np})
	\item NU: number of units. (\ref{nu})
	\item D: input dimension.
	\item NC: number of classes
\end{itemize}

We know that

\begin{equation}
	NP = D\times NU+NL(NU\times NC)+NL\times NU+NC
	\label{np}
\end{equation}

As a result, we can determine the number of neurons per layer:
\begin{equation}
	NU = \frac{NP-NC}{D+NL(NC+1)}
	\label{nu}
\end{equation}

\vspace{0.2cm}

Note that when addressing a regression problem, $NC=1$, for unidimensional output regression problems. It is straightforward to generalize this for multidimensional regression problems. As a result, we build a rectangular model whose complexity, in terms of the number of parameters, is enough to successfully complete the learning task.  

\begin{figure}[H]
	\centering
	\includegraphics[scale=0.35]{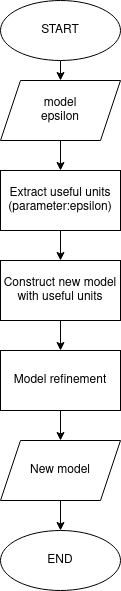}
	\caption{Algorithm 2. Flowchart}
	\label{alg2}
\end{figure}

After the aforementioned model (with NU units per layer) has been trained, based on \cite{ARTICULOMALIGNO}, we apply a pruning method which employs a three-step process (Algorithm 2, Figure \ref{alg2}). It receives two arguments: a feed-forward neural network, which has already been trained and whose predictions are accurate for a given problem, and a pruning tolerance parameter, $\epsilon > 0$.   \\

The process starts with the extraction of the \textit{useful units} or neurons from the model, in other words, those units which are indispensable for the network to acquire the real nature and patterns from data. 

\begin{algorithm}[H]
	\caption{Neural network refining procedure}
	\begin{algorithmic}
		\Function{RefiningNN}{model, $\epsilon$}
		\State Extract the useful units per layer from \textit{model} applying $\epsilon$.
		\State Construct the new model with the useful units and layers.
		\State Model refinement.
		\EndFunction
	\end{algorithmic}
\end{algorithm}

The second step is to construct a new model integrating all the layers, composed only of useful units, to evaluate the prediction capacities of the pruned neural network. The obtained model may under-perform the results compared to the original one. To overcome this problem, the pruned model is lightly retrained using a low number of epochs compared to the number of epochs employed by the training process of the original model. In consequence, the pruned model obtains competitive results and, in some cases, it may even outperform the original model.

\subsection{Extracting useful units}

The key idea behind the process that performs the extraction of useful units is to find the ``important'' neurons for the learning task. This process is completed to enhance the model by reducing its size and raising the accuracy level. The resulting model, due to its reduced size, will be more efficient, not only in terms of memory footprint but also runtime.

\begin{figure}[H]
	\centering
	\includegraphics[scale=0.35]{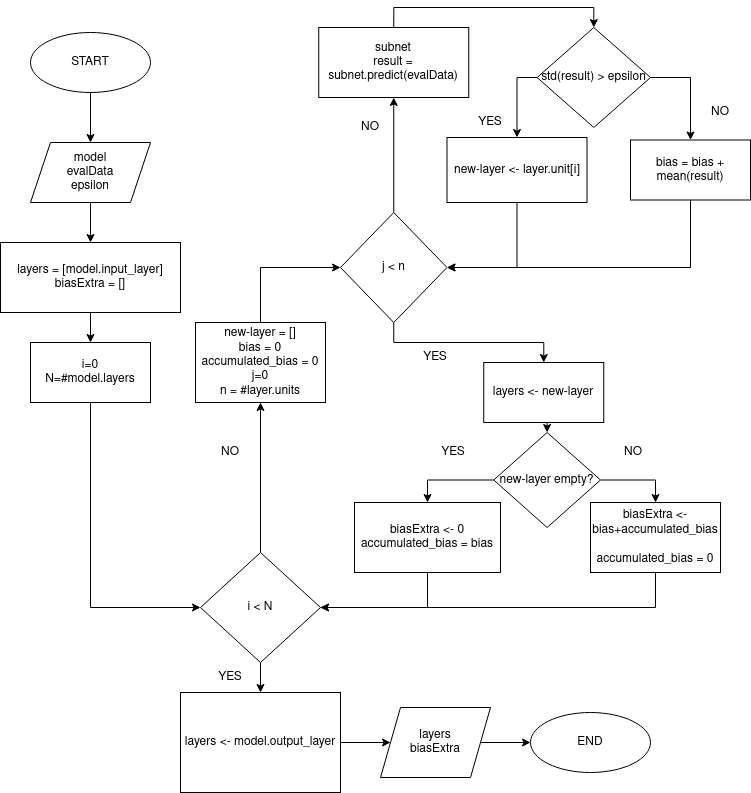}
	\caption{Algorithm 3. Flowchart}
	\label{alg3}
\end{figure}

Algorithm 3 (Figure \ref{alg3}) shows the extraction process if performed layer-wise examining each neuron. The inputs are a model, an evaluation dataset and a pruning tolerance level $\epsilon$.

\begin{algorithm}[]
	\caption{Extracting useful units}
	\begin{algorithmic}
		\Function{usefulUnits}{model, evalData, $\epsilon$}
		\State layers = []
		\State biasExtra = []
		\State Append to \textit{layers} the original input layer from \textit{model}
		\For{each hidden layer in \textit{model}, feed-forward wise,}
		\State new-layer = []
		\State bias = 0
		\State accumulated-bias = 0
		\For{each unit in layer}
		\State Create a subnet with $unit$ as its output layer
		\State result $\leftarrow$ Evaluate subnet on $evalData$
		\If{ Standard-deviation(\textit{result})  $> \epsilon$}
		\State Append the unit to \textit{new-layer}
		\Else
		\State \textit{bias} = \textit{bias} + mean(\textit{result})
		\EndIf
		\EndFor
		\State Append \textit{new-layer} to \textit{layers}
		\If{\textit{new-layer} is not empty }
		\State Append \textit{bias}+\textit{accumulated-bias} to \textit{biasExtra}
		\State \textit{accumulated-bias} = 0
		\Else
		\State Append 0 to \textit{biasExtra}
		\State \textit{accumulated-bias} = \textit{bias}
		\EndIf
		\EndFor
		\State Append to \textit{layers} the original output layer from \textit{model}
		\State \Return \textit{layers}, \textit{biasExtra}
		\EndFunction
	\end{algorithmic}
\end{algorithm}

\begin{figure}[H]
	\centering
	\includegraphics[scale=0.27]{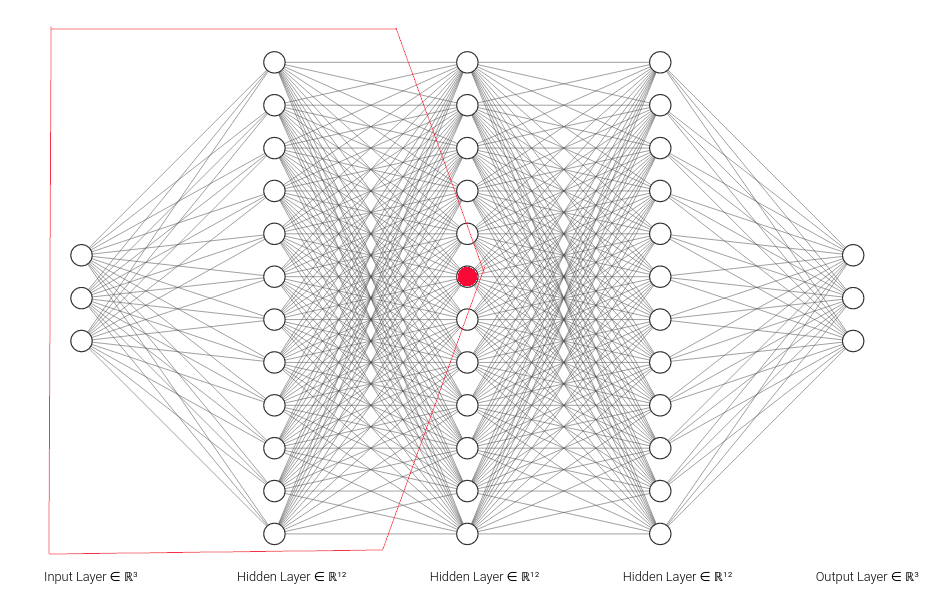}
	\caption{Implicit subnet for the candidate unit (red)}
	\label{red}
\end{figure}

To better expose the aforementioned algorithm, let us present an example. Let us consider the network shown in Figure \ref{red}. For the selected unit, its implicit subnet is bounded by the red pentagon.

Supposing that we are keen on evaluating the relevance of the unit in red, it is necessary to define a strategy to isolate the unit and its implicit subnet, in other words, the neural network formed by every unit belonging to the previous layers and whose output layer is the unit in red. 

Once we have a subnet which characterizes a neuron, we need to measure its importance within the whole model. The importance of a subnet is measured by its outputs, which are evaluated for different input examples (\textit{evalData}). If the output of the subnet varies sufficiently for different input examples, then the subnet is considered to contribute valuable insights to the overall information flow across the network. Conversely, if the output of the subnet is uniform for different input examples, then the neuron is deemed to be unimportant for the final prediction of the network. Thus, a metric is needed to encapsulate the variability of the output of a subnet. The standard deviation is chosen as the metric due to its simplicity and widespread use in statistics. The standard deviation over the \textit{evalData} outputs' is measured and if it is greater than $\epsilon$, the unit is classified as significant (useful unit). In consequence, it will form part of the refined model. Otherwise, this unit will not appear in the new model. 

However, even if the variation of the predictions in a given subnet might be subtle, it could have an important and global effect on the neural network. Therefore, these results should be taken into consideration. Bearing that in mind, for those units that will be discarded in the new model, we add the mean of their outputs to the layer's bias. As a result, we take into account all the information from the neurons but we only compute those which produce enough variability. \\

\begin{figure}[H]
	\centering
	\includegraphics[scale=0.4]{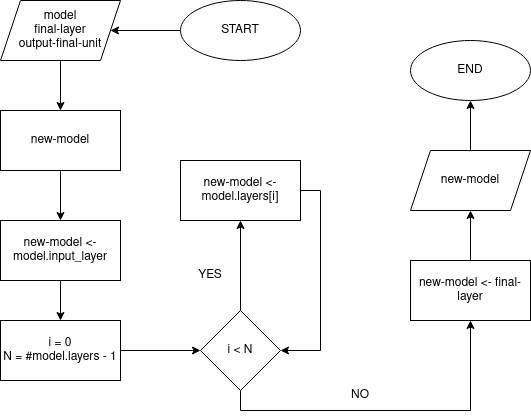}
	\caption{Algorithm 4. Flowchart}
	\label{alg4}
\end{figure}

In Algorithm 4 (Figure \ref{alg4}), we can see how a subnet is built using a specific neuron as the only output unit. First, we create a new model and add the input layer from the original model. Then, we add to the new model all the previous layers before the one which contains the neuron being considered for evaluation. Once we find the neuron, we add a layer with this single neuron as the output layer to the subnet. No possible conflict arises in the construction since the \textit{useful units} are selected feed-forward wise.

\begin{algorithm}[H]
	\caption{Creating a subnet with a specific output final unit}
	\begin{algorithmic}
		\Function{SubNet}{model, final-layer, final-unit}
		\State Create a new model (\textit{new-model})
		\State Add the input layer from the original \textit{model}
		\For{each layer in \textit{model} previous to \textit{final-layer}}
		\State Add \textit{layer} to \textit{new-model} and set weights to \textit{new-model}
		\EndFor
		\State Add a final layer with an unique output unit (\textit{final-unit} from \textit{final-layer}) to \textit{new-model}
		\State \Return \textit{new-model}
		\EndFunction
	\end{algorithmic}
\end{algorithm}

Each of these sub networks gather a part of the knowledge captured by the initial large network. Actually, a rather relevant part of the overall knowledge since it has been selected as useful. By reusing this part of the network the user is achieving knowledge transfer.

\subsection{Building and retraining the refined model}

Once we have defined the useful unit list, we build the refined model using those units which are in the list (whose outputs present enough variability) and adding them to the new network respecting the original topology related to the useful units. Careful update of the weights and shape is crucial in this phase, due to the fact that our method may detect the inoperability of a complete layer in the original model and thus, the layer would be completely disregarded in the process of building the refined model.  In this case, if the $n$th layer is discarded, the ($n-1$)th and the ($n+1$)th layers will be connected. Therefore, an adjustment of shapes is needed. The only layers which are kept immutable are the input and the output layers. Hidden layers from the original model may change completely. 

However, as we have said above, the refined model may underperform the results compared to the original one. Therefore, a retraining phase (model refinement) is necessary to finish the process. We have tried to retrain as less as possible so as to carry out small adjustments and avoid overfitting. In other words, we set the number of epochs to 15\% of the original number of epochs at the beginning of the process. Hyper parameters like activation functions, optimizer or batch size are maintained constant in both learning tasks. 

Our method could be at first considered only as an effective way to only find a neural network optimal topology for a specific learning task. Nonetheless, we have found that building a network with the topology identified by our algorithm and then setting random initial weights and training (with any of the currently available learning algorithms) is not enough to reach the same performance as that achieved with the refined model. Therefore, our method is not only guiding us in the selection of the neural network topology but also it is transferring knowledge from the original model to the refined one. In short, our methodology implements a particular kind of transfer learning. Thus, we could see this method simply as a neural network design procedure characterized by transfer learning.

\section{Empirical Evaluation}

We have thoroughly evaluated our neural network building scheme, considering the two broad classes of problems —classification and regression—, detailing the study throughout diverse categories of classification categories. ODF2NNA has been compared with the principal state-of-art approaches mentioned in Section II using the datasets proposed in the research papers which presented those techniques.  Additionally, two additional standard classification problems (Spambase and Poker) and one large-scale dataset (Hepmass) from the UCI Machine Learning Repository have been utilized to evaluate the performance and quality of our dense neural network optimization algorithm. Finally, we consider three regression datasets (Ailerons, Compactiv and Pole) to evaluate the effectiveness of our method not only in classification but also in regression. In each experiment, several values for $\epsilon$ were used to evaluate the learning capacity of the refined model obtained and its effectiveness. We measured the prediction performance with accuracy (ACC) metrics for classification and mean squared error (MSE) for regression. In addition, we registered the number of parameters and the number of layers of the model to assess the efficiency in terms of memory requirements and runtime (the lower the number of parameters, the higher the efficiency). We compared the accuracy (or MSE, respectively) in three different stages: original model, pruned model without retraining and refined model (retrained pruned model). As a result, we are able to understand if the original model was too complex. In other words, if the original model had more parameters than actually needed. Finally, we can notice the pruned model's learning capacity by comparing the accuracy (or MSE) before and after retraining. The process of retraining is very light in terms of epochs (10\% - 15\% of the original number or epochs) and, thus, we avoid overfitting. 

In summary, we compared ODF2NNA with 15 different techniques for pruning feed-forward neural networks. Additionally, we extended the experimentation to problems that have not been addressed, to the best of our knowledge, in the literature, such as dividing the experimentation according to the size of the dataset (medium-scale datasets, large-scale datasets) and facing regression problems.

We implemented the proposal using Keras-Tensorflow 2.1. We carried out the experiments on the following pieces of hardware: AMD Ryzen 7 3800X 8-Core Processor, NVIDIA GeForce GTX 1650 SUPER and NVIDIA GeForce RTX 2080 ti. 

\subsection{Classification experiments}

To illustrate the generality of our method we test it on a core set of common benchmark datasets and fully connected network architectures. In all of these experiments, we measure the accuracy (or the classification error rate) before and after applying the optimization method and the complexity reduction in terms of the remaining number of parameters or layers. Nonetheless, we have adapted our metrics (and the way we show them) to the ones used in the state-of-art references to achieve an accurate comparison between ODF2NNA and the state-of-art references.

\subsubsection{Experiment 1: Well-known datasets}

In this section, the proposed algorithm is evaluated on four well known datasets from UCI Machine Learning Repository and compared to other pruning methods such as VNP \cite{VNP}, Xing-Hu's method \cite{Xing2009}, MBP \cite{HAGIWARA1994207}, OBD \cite{OBD}, OBS \cite{OBS} and N2PS \cite{Augasta2011ANP}. The datasets used in this experiment are:

\begin{itemize}
	\item Iris dataset (iris): 150 instances, four attributes and three categories: setosa, versicolor and virginica.
	\item Wisconsin-breast-cancer dataset (cancer): It contains 699 instances with 9 real value attributes and two classes.
	\item Hepatitis domain dataset (hepatitis): 155 instances, 19 attributes. Binary classification.
	\item Pima Indians Diabetes dataset (diabetes): Binary classification problem with 500 instances and 8 attributes.
\end{itemize}

Table \ref{table1} shows the comparison results of our method on the four datasets compared to the other pruning methods:
\begin{table}[H]
	\begin{adjustbox}{width=\textwidth}
		\begin{tabular}{|c|c|c|c|c|c|c|c|c|c|c|c|c|c|c|c|c|}
			\hline
			Dataset   & Original &      & N2PS  &       & VNP   &      & Xing-Hu &       & OBD    &      & OBS     &      & MBP     &      & \begin{tabular}[c]{@{}c@{}}Our\\  approach\end{tabular} &                \\ \hline
			& NN       & ACC  & NN    & ACC   & NN    & ACC  & NN      & ACC   & NN     & ACC  & NN      & ACC  & NN      & ACC  & NN                                                      & ACC            \\ \hline
			Iris      & 4-10-3   & 96   & 3-3-3 & 98.67 & 2-2-3 & 97.7 & 3-2-3   & \textbf{98.67} & 4-4-3  & 98   & 4-4-3   & 98   & 4-4-3   & 98   & 3-4-3                                                   & 98.5           \\ \hline
			Cancer    & 9-10-2   & 95.4 & 3-2-2 & 97.1  & 9-1-2 & 97.8 & 3-3-2   & 96.78 & 9-8-1  & 92.5 & 9-7-1   & 90   & 9-7-1   & 90   & 5-3-2                                        & \textbf{97.86} \\ \hline
			Hepatitis & 19-25-2  & 80.2 & 2-3-2 & 86.4  & 4-4.2 & 83.3 & 3-8-2   & 84.62 & 19-9-1 & 78.7 & 19-16-1 & 73.8 & 19-18-1 & 80.3 & 7-11-2                                                  & \textbf{87.21} \\ \hline
			Diabetes  & 8-40-2   & 68.6 & 5-3-2 & 70.3  & 6-8-2 & 69.1 & 6-8-2   & \textbf{74.22} & 8-16-1 & 68.6 & 8-26-1  & 65.4 & 8-26-1  & 68.9 & 5-17-2                                                  & 73.62 \\ \hline
		\end{tabular}
	\end{adjustbox}
	\caption{Comparison results of ODF2NNA on iris, cancer, hepatitis and diabetes. NN: Neural Network Architecture ($<\#$units layer 1$>$-$<\#$units layer 2$> \dots <\#$units layer $n>$)}. Best accuracy results are highlighted in bold.
	\label{table1}
\end{table}

As we can see, ODF2NNA outperforms the accuracy results for Cancer and Hepatitis compared to the other state-of-art methods reducing the architecture widely.

Table \ref{table1-f1} shows the F1-Score obtained for Iris, Wisconsin-breast-cancer, Hepatitis and Diabetes datasets. We cannot compare F1-Score with the other state-of-art methods because it is not included in their results.
% IRIS F1-SCORE 0.98
% Cancer f1-score 0.86
% Hepatitis f1-score 0.79
% Diabetes f1-score 0.66

\begin{table}[H]
	\centering
	\begin{tabular}{|c|c|c|c|c|}
		\hline
		& Iris & Cancer & Hepatitis & Diabetes \\ \hline
		F1-Score & 0.98 & 0.86   & 0.79      & 0.66     \\ \hline
	\end{tabular}
	\caption{F1-Score value}
	\label{table1-f1}
\end{table}

\subsubsection{Experiment 2: Modifying the number of layers}

We tested our method on deep neural networks with different numbers of hidden layers (from 2 to 6) on the MNIST dataset. The MNIST dataset consists of $28 \times 28$ pixel handwritten digit images from 0 to 9. The task is to classify the images into ten digit classes. There are 60 000 examples in the training set and 10 000 in the test set. For all models, we allocate 1000 units and compare the results with the Learning Automata-Based SGD \cite{7939970}. Table \ref{table2} shows the classification error percentage with different number of hidden layers (columns 2, 3 and 4). The Sparseness column denotes the rates of remaining weights after pruning with respect to the initial number of weights (less is better).
F1-Score value is also added. As mentioned before, we cannot compare F1-Score with the other state-of-art methods because it is not included in their investigation results.

\begin{table}[H]
	\begin{adjustbox}{width=\textwidth}
		\begin{tabular}{|c|c|c|c|c|c|c|}
			\hline
			Hidden layers & SGD  & SGD+LA & ODF2NNA    & Sparseness - Guo et al. & ODF2NNA sparseness & ODF2NNA F1-Score \\ \hline
			$1000\times2$        & 1.45 & 1.43   & 1.87          & 43\%                    & \textbf{5\%}  &  0.79 \\ \hline
			$1000\times3$        & 1.56 & 1.52   & \textbf{1.49} & 62\%                    & \textbf{32\%} &  0.94\\ \hline
			$1000\times4$        & 1.61 & 1.49   & 1.51          & 60\%                    & \textbf{27\%} &  0.81\\ \hline
			$1000\times5$        & 1.79 & 1.47   & 1.68          & 44\%                    & \textbf{13\%} &  0.84\\ \hline
			$1000\times6$        & 2.4  & 1.58   & \textbf{1.52}           & 54\%                    & \textbf{19.8\%}& 0.87 \\ \hline
		\end{tabular}
	\end{adjustbox}
	\caption{Classification error and sparseness. Best results are highlighted in bold.}
	\label{table2}
\end{table}

The least error measure values obtained are for $1000\times3$ and $1000\times6$ hidden layers. Besides, we are able to reduce the sparseness in each architecture compared to Guo et al. \cite{7939970}.

\subsubsection{Experiment 3: LeNet300-100}

In this experiment we verify the effectiveness of our optimization method on MNIST using a deep neural network architecture called LeNet-300-100, which is a fully connected neural network made of four layers: the input layer, the output layer and two hidden layers with 300 and 100 neurons respectively. 

First of all, we compare our method with Han et al. \cite{han2016deep} and Manessi et al. \cite{8546129}. Table \ref{table3} shows the percentage prediction error ($\Delta_1$) for the pruned networks  and their respective non-pruned implementation. We also show the pruning performance, measured by the number of weights' compression. Similarly, Han et al.  \cite{han2016deep} and Manessi et al. \cite{8546129} results are considered. Note that there are N/A values, which mean that the original research paper did not specified any value for this item.

\begin{table}[H]
	\begin{tabular}{|l|c|c|c|}
		\hline
		\hspace{2.5cm}Method                                & $\Delta_1$  & Weights & Pruning Performance \\ \hline
		LeNet-300-100 reference               & N/A           & 266K    & N/A       \\ \hline
		LeNet-300-100 pruned (Han et al.)     & 0.1\%            & 21K     & 12x     \\ \hline
		LeNet-300-100 pruned (Manessi et al.) & 0.1\%            & 14K     & 19x     \\ \hline
		LeNet-300-100 pruned (ODF2NNA)   & 0.2\%           & \textbf{5K}      & \textbf{40x}     \\ \hline
	\end{tabular}
	\caption{LeNet-300-100 (I) Comparison with  \cite{han2016deep} and  \cite{8546129}. Best results are highlighted in bold.}
	\label{table3}
\end{table}

As we can see, the number of weights is widely reduced, improving the pruning performance.

Finally, we compare our method with randomly pruning \cite{dong2017learning}, Layer-wise OBS \cite{dong2017learning},  OBD \cite{OBD}, LWC \cite{LWC} and DNS \cite{guo2016dynamic}. In this case, we contemplate the original classification error, the error rate after pruning, the error rate after retraining (Re-Error), the number of iterations to complete the retraining phase (\#Re-Iters) and the compression ratio (CR), which is the ratio of the number of preserved parameters to that of original parameters (lower is better) \cite{dong2017learning}. Table \ref{table4} shows the obtained results:

\begin{table}[H]
	\begin{adjustbox}{width=\textwidth}
		\begin{tabular}{|c|c|c|c|c|c|}
			\hline
			Method            & Original Error & CR     & Err. After Pruning & Re-Error        & \#Re-Iters  \\ \hline
			Random            & 1.76\%         & 8\%    & 85.72\%            & 2.25\%          & 3.5e5       \\ \hline
			OBD               & 2.76\%         & 8\%    & 86.72\%            & 1.96\%          & 8.1e4       \\ \hline
			LWC               & 3.76\%         & 8\%    & 81.32\%            & 1.95\%          & 1.4e5       \\ \hline
			DNS               & 4.76\%         & 1.8\%  & N/A                  & 1.99\%          & 3.4e4       \\ \hline
			L-OBS             & 5.76\%         & 7\%    & 3.1\%              & 1.82\%          & 510         \\ \hline
			L-OBS (iterative) & 6.76\%         & 1.55\% & 2.43\%             & 1.96\%          & 643         \\ \hline
			ODF2NNA      & 2\%            & 2.46\% & 92\%               & \textbf{1.78\%} & \textbf{50} \\ \hline
		\end{tabular}
	\end{adjustbox}
	\caption{LeNet-300-100 (II). Comparison with \cite{dong2017learning}, \cite{OBD}, \cite{LWC} and \cite{guo2016dynamic}. Best results are highlighted in bold.}
	\label{table4}
\end{table}

Even if ODF2NNA's compression rate is higher than L-OBS (iterative) and DNS, the error rate after the retraining phase is lower. In addition, a significant difference can be observed in the number of epochs required by ODF2NNA to fine-tune the pruned optimized network (50) compared to its closest competitor in error rate after retraining, L-OBS (iterative), which required 643 retraining epochs. This indicates not only that ODF2NNA is a more effective method for optimizing the original network but also that it is much more efficient in terms of runtime and energy consumption. Additionally, F1-Score obtained is 0.77. Again, we cannot compare it with the other state-of-art methods because it is not included in their investigation results.

\subsubsection{Experiment 4: Other UCI datasets}

In order to complete our empirical evaluation, we chose two extra datasets from the UCI Machine Learning Repository which are more complex than the datasets used in Experiment 1. These datasets are Spambase (4597 examples, 57 real attributes, and binary classification) and Poker (1025010, 10 integer attributes, multi classification). To the best of our knowledge, these datasets have not been used to evaluate dense neural network optimization algorithms. Table \ref{table5} (Spambase) and Table \ref{table6} (Poker) show the original accuracy rate (OC), the accuracy rate after pruning without retraining (PWR) and the accuracy rate after retraining (PR). Each table shows the original number of parameters (OR), the number of preserved parameters after pruning (P) and the reduction percentage.

\begin{table}[H]
	\centering
	\begin{tabular}{|c|c|c|c|c|c|c|}
		\hline
		& \multicolumn{3}{c|}{\textbf{ACCURACY}} & \multicolumn{3}{c|}{\textbf{N. PARAM}} \\ \hline
		$\epsilon$ & OR          & PWR         & PR         & OR        & P       & \% Reduc     \\ \hline
		0.1        & 75\%        & 76\%        & 91\%       & 1262      & 993     & 21.32\%          \\ \hline
		0.3        & 75\%        & 77\%        & 93\%       & 1262      & 861     & 31.77\%          \\ \hline
		0.5        & 75\%        & 39\%        & 93\%       & 1262      & 836     & 33.76\%          \\ \hline
		0.7        & 75\%        & 39\%        & 93\%       & 1262      & 824     & 34.71\%          \\ \hline
		0.9        & 75\%        & 39\%        & 92\%       & 1262      & 810     & 35.82\%          \\ \hline
	\end{tabular}
	\caption{Spambase dataset results (OR = original, PWR = Pruning without retraining, PR = Pruned and retrained, P = pruned)}
	\label{table5}
\end{table}

\begin{table}[H]
	\centering
	\begin{tabular}{|c|c|c|c|c|c|c|}
		\hline
		& \multicolumn{3}{c|}{\textbf{ACCURACY}} & \multicolumn{3}{c|}{\textbf{N. PARAM}} \\ \hline
		$\epsilon$ & OR          & PWR         & PR         & OR        & P        & \% Reduc    \\ \hline
		0.1        & 53\%        & 50\%        & 59\%       & 374261    & 137141    & 63.36\%         \\ \hline
		0.3        & 53\%        & 42\%         & 66\%       & 374261    & 55791    & 85.09\%         \\ \hline
		0.5        & 53\%        & 0\%        & 74\%       & 374261    & 7     21776 & 94.18\%         \\ \hline
		0.7        & 53\%        & 0\%        & 41\%       & 374261    &
		9477     & 97.47\%         \\ \hline
		0.9        & 53\%        & 0\%        & 86\%       & 374261    & 3837     & 98.97\%         \\ \hline
	\end{tabular}
	\caption{Poker dataset results (OR = original, PWR = Pruning without retraining, PR = Pruned and retrained, P = pruned)}
	\label{table6}
\end{table}

It should be remarked that the pruned and retrained model produces more accurate predictions than the original model in every case. Besides, the reduction in the number of parameters is enormous for the Poker dataset. The best Spambase F1-Score obtained is 0.89 ($\epsilon = 0.7$) and the best Poker F1-Score obtained is 0.65 ($\epsilon=0.5$).

\subsubsection{Experiment 5: Large-scale dataset}

The experiments above show that our method can be successfully applied to small and medium-scale datasets and deep neural architectures. After thoroughly reviewing the literature, we have not found any fully connected neural network optimization method evaluated on large-scale datasets. We have selected Hepmass from the UCI Machine Learning Repository to verify that ODF2NNA is also effective for large-scale datasets (10,500,000 instances, 28 attributes).

\begin{table}[H]
	\begin{adjustbox}{width=\textwidth}
		\begin{tabular}{|c|c|c|c|c|c|}
			\hline
			$\epsilon$ & \#Layers & \# Parameters & \#Epochs train/retrain & Accuracy & \%Reduction \\ \hline
			Original   & 17       & 3773002       & 1000                   & 81\%     & 0\%           \\ \hline
			5e-5       & 17       & 2287537       & 100                    & 82\%     & 39.37\%     \\ \hline
			1e-5       & 17       & 2283005       & 100                    & \textbf{85.6}\%   & \textbf{39.49}\%     \\ \hline
			0.0001     & 17       & 2224246       & 100                    & 83.49\%  & 41.05\%     \\ \hline
			0.0004     & 17       & 2114143       & 100                    & 80.35\%  & 43.96\%     \\ \hline
			0.01       & 17       & 1274437       & 100                    & 76.47\%  & 66.22\%     \\ \hline
			0.04       & 13       & 597337        & 100                    & 78.51\%  & 84.17\%     \\ \hline
			0.2        & 10       & 396290        & 100                    & 56.21\%  & 89.5\%      \\ \hline
			0.8        & 8        & 257858        & 100                    & 53.9\%   & 93.17\%     \\ \hline
			1          & 7        & 230417        & 100                    & 50\%     & 93.89\%     \\ \hline
		\end{tabular}
	\end{adjustbox}
	\caption{Hepmass dataset results}
\end{table}

As we can see, the value $\epsilon = 1e-5$ produces the best results, including F1-Score (0.72).\\

In addition, we have selected Higgs from the UCI Machine Learning Repository as another example of learning task based on a large dataset (11,000,000 instances, 28 attributes). This dataset size is comparable in terms of instance number to the widely considered Imagenet benchmark. The Higgs dataset was used in \cite{Baldi2014}, as a semi-realistic case, for showing that deep learning can improve the power of collider searches for exotic particles, concluding that deep-learning techniques discover powerful nonlinear feature combinations and provide better discrimination power than other classifiers.

In \cite{Baldi2014} a feed-forward network architecture is presented, formed by five layers with 300 hidden units in each layer, a learning rate of $0.05$ and a weight decay of $1e-5$, obtaining an area under the curve (AUC) value of $0.885$. We have replicated this architecture and conditions to apply ODF2NNA, with a retraining phase of 50 epochs.

\begin{table}[H]
	\begin{adjustbox}{width=\textwidth}
		\begin{tabular}{|c|c|c|c|c|c|}
			\hline
			$\epsilon$ & \#Layers & \# Parameters & \#Epochs train/retrain & Accuracy & \%Reduction \\ \hline
			Original   & 5       & 370502       & 1000                   & 88\%     & 0\%           \\ \hline
			$0.0001$       & 5       & 283895       & 50                    & 91.3\%     & 23.38\%     \\ \hline
			$0.0002$       & 17       & 282055       & 50                    & \textbf{93}\%   & \textbf{23.87}\%     \\ \hline
			$0.0003$     & 17       & 280309       & 100                    & 89\%  & 24.34\%     \\ \hline
			$0.0004$     & 17       & 247669       & 100                    & 83.6\%  & 33.15\%     \\ \hline
			$0.001$       & 13       & 192351        & 100                    & 76.13\%  & 48.08\%     \\ \hline
			$0.01$        & 10       & 95363        & 100                    & 69.87\%  & 74.26\%      \\ \hline
		\end{tabular}
	\end{adjustbox}
	\caption{The Higgs dataset results}
\end{table}

\begin{figure}[H]
	\centering
	\includegraphics[scale=0.5]{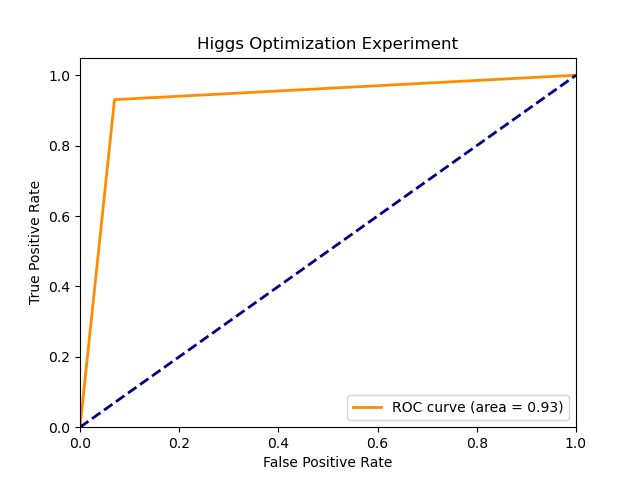}
	\caption{ROC Curve for the Higgs Dataset optimization $\epsilon=0.0002$}
	\label{roc_higgs}
\end{figure}

ODF2NNA is able to reduce by more than 23\% the number of parameters and increases in a 5\% the accuracy compared to the original model. As we can see in Figure \ref{roc_higgs}, the AUC for the best $\epsilon$ is $0.93$, compared to the $0.885$ reported in \cite{Baldi2014}. Again, it is clear that ODF2NNA is an accurate optimization method for learning tasks based on large datasets.

\subsection{Regression}

We evaluated ODF2NNA not only on classification problems but also on regression ones. For this reason, we chose three regression datasets from UCI Machine Learning Repository: Ailerons, Compactive and Pole. The result tables of this section present the values of Mean Squared Error (MSE) and the number of parameters (N.PARAM) for each dataset.

\begin{table}[H]
	\centering
	\begin{tabular}{|c|c|c|c|c|c|c|}
		\hline
		& \multicolumn{3}{c|}{\textbf{MSE}} & \multicolumn{3}{c|}{\textbf{N. PARAM}} \\ \hline
		$\epsilon$ & OR        & PWR       & PR        & OR       & P        & \% Reduc     \\ \hline
		0.002      & $3.07 e^{-6}$ & $4.06 e^{-5}$ & $3.28 e^{-6}$ & 2809     & 1148     & 59.13\%          \\ \hline
		0.008      & $3.07 e^{-6}$ & $0.0009$  & $1.42 e^{-5}$ & 2809     & 771      & 72.55\%          \\ \hline
		0.014      & $3.07 e^{-6}$ & $6.68 e^{-5}$ & $2.64 e^{-6}$ & 2809     & 757      & 73.05\%          \\ \hline
		0.020       & $3.07 e^{-6}$ & $2.66 e^{-5}$ & $2.74 e^{-6}$ & 2809     & 757      & 73.05\%          \\ \hline
		0.026      & $3.07e^{-6}$ & $2.65e^{-5}$ & $3.04 e^{-6}$ & 2809     & 757      & 73.05\%          \\ \hline
	\end{tabular}
	\caption{Ailerons dataset results (OR = original, PWR = Pruning without retraining, PR = Pruned and retrained, P = pruned)}
\end{table}

\begin{table}[H]
	\centering
	\begin{tabular}{|c|c|c|c|c|c|c|}
		\hline
		& \multicolumn{3}{c|}{\textbf{MSE}}                                     & \multicolumn{3}{c|}{\textbf{N. PARAM}} \\ \hline
		$\epsilon$ & OR                    & PWR                   & PR                    & OR       & P        & \% Reduc    \\ \hline
		0.002      & $2.63e^{-7}$ & $2.19e^{-6}$ & $1.24e^{-6}$ & 1958     & 1279     & 34.68\%          \\ \hline
		0.008      & $2.63e^{-7}$ & $1.14e^{-5}$ & $4.81e^{-6}$ & 1958     & 513      & 73.8\%           \\ \hline
		0.014      & $2.63e^{-7}$ & $3.86e^{-5}$ & $5.55e^{-7}$ & 1958     & 438      & 77.63\%          \\ \hline
		0.020       & $2.63e^{-7}$ & $3.86e^{-5}$ & $6.38e^{-7}$ & 1958     & 438      & 77.63\%          \\ \hline
		0.026      & $2.63e^{-7}$ & $3.86e^{-5}$ & $8.61e^{-7}$  & 1958     & 438      & 77.63\%          \\ \hline
	\end{tabular}
	\caption{Compactiv dataset results (OR = original, PWR = Pruning without retraining, PR = Pruned and retrained, P = pruned)}
\end{table}

\begin{table}[H]
	\centering
	\begin{tabular}{|c|c|c|c|c|c|c|}
		\hline
		& \multicolumn{3}{c|}{\textbf{MSE}}                                     & \multicolumn{3}{c|}{\textbf{N. PARAM}} \\ \hline
		$\epsilon$ & OR                    & PWR                   & PR                    & OR        & P       & \% Reduc     \\ \hline
		0.002      & $6.35e^{-5}$ & $7.43e^{-5}$ & $1.76e^{-6}$ & 4159      & 973     & 76.6\%           \\ \hline
		0.008      & $6.35e^{-5}$ & $5.64e^{-5}$ & $5.66e^{-6}$ & 4159      & 815     & 80.4\%           \\ \hline
		0.014      & $6.35e^{-5}$ & $5.94e^{-5}$ & $7.05e^{-6}$ & 4159      & 650     & 84.37\%          \\ \hline
		0.020       & $6.35e^{-5}$ & $5.72e^{-5}$ & $7.35e^{-6}$ & 4159      & 617     & 85.16\%          \\ \hline
		0.026      & $6.35e^{-5}$ & $7.32e^{-5}$ & $6.87e^{-6}$ & 4159      & 540     & 87.02\%          \\ \hline
	\end{tabular}
	\caption{Pole dataset results (OR = original, PWR = Pruning without retraining, PR = Pruned and retrained, P = pruned)}
\end{table}

It should be remarked that a high reduction in the number of parameters is achieved without a relevant increase in the error measure.
\section{Empirical result analysis and Discussion}

We start this section analyzing the classification results. The classification experiments are divided into five sections, where the size of the datasets is varied to verify the robustness of ODF2NNA when facing different learning tasks. In total, the results of ODF2NNA are compared with 13 techniques from the state of the art. The following presents the different proposed experiments.

Experiment 1 was designed to evaluate ODF2NNA on small-scale datasets. We compared its performance to other pruning methods which were built to face this kind of learning tasks. We found that ODF2NNA outperforms all the other approaches, in terms of neural network complexity (NN) and accuracy for the Cancer dataset. For Hepatitis and Diabetes, we obtained the highest accuracy after optimizing the model.

Experiment 2 measured the classification error after retraining and the optimized model sparseness, in other words, the rates of remaining weights after pruning. Using the same number of hidden layers (from 2 to 6) as the original work \cite{guo2016}, ODF2NNA obtained very similar classification error rates while generating a model whose complexity in terms of the number of parameters was reduced by more than 60\% compared to \cite{guo2016}. In fact, for two hidden layers, the rate of remaining weights after pruning dropped to one tenth (43\% vs 5\%); for three and four hidden layers, around half as much (62\%-32\%, 60\%-27\%); for five hidden layers, it dropped to three tenth (44\%-13\%); and for six hidden layers, the number of remaining weights dropped to one third (54\%-19.8\%). In consequence, in this test ODF2NNA showed an outstanding capacity of reducing deep learning models and robustness even when the neural network architecture is heavily modified. Experiment 3 endorsed this statement too, since we managed to reduce the LeNet 300-100's number of parameters from 266000 to 5000, in other words, the amount of parameters dropped by 98\% without worsening the performance in prediction. Experiment 3 also showed that ODF2NNA is effective not only finding a new optimized model from an original one given a learning task, but also highly efficient due to the low amount of iterations to complete the retraining phase (\#Re-Iters in Table \ref{table4}) compared to other approaches. 

\begin{figure}[H]
	\begin{subfigure}{0.5\textwidth}
		\includegraphics[width=0.9\linewidth, height=5cm]{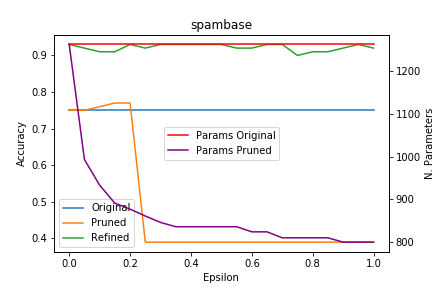} 
		\caption{Spambase}
		\label{fig:spambase}
	\end{subfigure}
	\begin{subfigure}{0.5\textwidth}
		\includegraphics[width=0.9\linewidth, height=5cm]{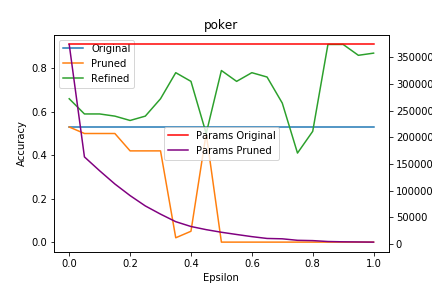} 
		\caption{Poker}
		\label{fig:poker}
	\end{subfigure}
	\caption{Accuracy performance vs. N. Parameters}
	\label{alwayswin}
\end{figure}

Experiment 4 showed that Spambase (Figure \ref{fig:spambase}) and Poker (Figure \ref{fig:poker}) are two examples which illustrate how effective pruning and retraining a model can be, increasing by more than 15\%, in the case of Spambase, and by more than 40\% for certain values of $\epsilon$, in the case of Poker, the accuracy with respect to the original model. Besides, we reduced the number of parameters by more than 35\% for Spambase. As a result, we obtained, for all $\epsilon \in (0, 1]$, a much more effective and efficient refined model to solve the Spambase learning task compared to the original model. For Poker, ODF2NNA reduced the number of parameters by more than 90\%. Nonetheless, for $\epsilon=0.7$, the refined model underperforms the original one, in other words, there are some $\epsilon$ values which generate a refined model less accurate than the original model. Therefore, it is clear that $\epsilon$ ought to be carefully chosen.

We finished the classification phase with Experiment 5, in which we faced two large-scale learning tasks: Hepmass and Higgs classification problems. In the case of Hepmass, the original model, which had more than 3.7 million of parameters, obtained an 81\% of accuracy in prediction. We found that, for $\epsilon = 1e-5$, the number of parameters was reduced by almost 40\% (2.28 million of parameters) and the accuracy was increased by more than 4.5\% (85.6\% in test). For the Higgs dataset, we used the architecture proposed in \cite{Baldi2014} in order to compare the ODF2NNA results and the original ones. A 93\% of accuracy in prediction, reducing in more than 23\% the number of parameters, and a $0.93$ in AUC is obtained, compared to the $0.885$ that can be found in the original paper. In other words, ODF2NNA is effective in large-scale learning tasks too. 

On the other hand, regression results are similar to classification ones. We can notice the importance of retraining the model again, obtaining in almost every case smaller mean square error values if a light retraining is applied. 

\begin{figure}[H]
	\begin{subfigure}{0.5\textwidth}
		\includegraphics[width=0.9\linewidth, height=5cm]{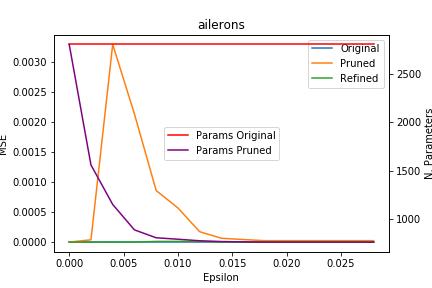} 
		\caption{Ailerons}
		\label{fig:ailerons}
	\end{subfigure}
	\begin{subfigure}{0.5\textwidth}
		\includegraphics[width=0.9\linewidth, height=5cm]{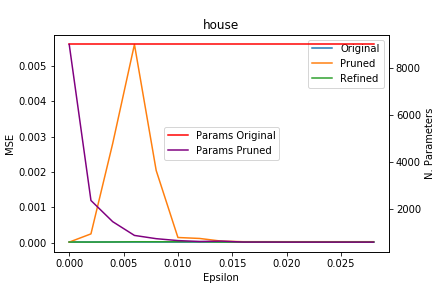}
		\caption{House}
		\label{fig:house}
	\end{subfigure}
	\caption{MSE performance vs. N. Parameters}
	\label{fig:regeq}
\end{figure}

\begin{figure}[H]
	\begin{subfigure}{0.5\textwidth}
		\includegraphics[width=0.9\linewidth, height=5cm]{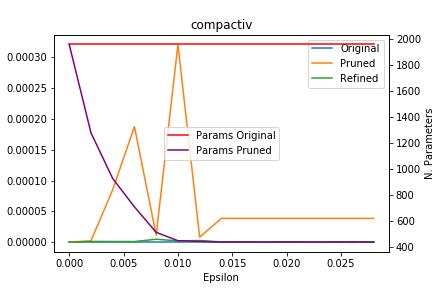} 
		\caption{Compactiv}
		\label{fig:compactiv}
	\end{subfigure}
	\begin{subfigure}{0.5\textwidth}
		\includegraphics[width=0.9\linewidth, height=5cm]{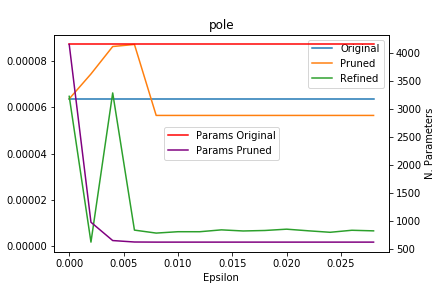}
		\caption{Pole}
		\label{fig:pole}
	\end{subfigure}
	\caption{MSE performance vs. N. Parameters}
	\label{fig:better}
\end{figure}

The complexity of the original models for regression problems is reasonable (they are not oversized), taking into account their acceptable original performance in contrast to the poor MSE results obtained after the simplification procedure without retraining. This can be considered in the classification examples given above, too. We found many examples where our construction method generates models with less than 70\% of the original number of parameters without affecting their performance in prediction. That is the case of Ailerons (Figure \ref{fig:ailerons}), House (Figure \ref{fig:house})  and Compactiv (Figure \ref{fig:compactiv}).

Again, we emphasize the necessity of carefully choosing the value of $\epsilon$. In our sensitivity study we can see again that, for some values of the pruning parameter, it is possible not only to maintain the original MSE in the refined model but also to improve it. That is the case of $\epsilon=0.01$ in Pole (MSE value of $7\text{·}10^{-6}$ versus $6.35\text{·}10^{-5}$ and 80\% of reduction of the number of parameters) (Figure \ref{fig:pole}).

Finally, it could seem reasonable to think that the refining method could be used only to find the optimal topology, namely, the number of layers and unit distribution over them (most challenging parameters for the data scientist), and that a model with this topology built from scratch would reach the same performance as the refined model. However, our experiments show that this is not true at least with the currently available training algorithms. 

We have selected Spambase (classification) and Pole (regression), two datasets where, for certain values of $\epsilon$, the refined method outperforms the original one.  In the case of Spambase, the optimal number of parameters after pruning is around 800 (35.82\% reduction compared to the original model). We have built a neural network with 800 parameters, Adam optimizer and 240 epochs and early stopping with patience of 20 (the same hyper parameters as the original model). We obtained 75\% of accuracy in prediction, which is smaller than the refined one (92\%). On the other hand, for Pole, we built a model with 815 parameters (the optimal number of parameters as we can see in the above section), Adam optimizer and 1000 epochs and early stopping with patience of 50 (same parameters as the original model) obtaining a mean squared error of $6.35\text{·}10^{-5}$, more than ten times greater than the MSE attained with the refined model ($5.66\text{·}10^{-6}$). Similar results have been obtained on other classification and regression problems.

As a result, our refined method is not only guiding us in the construction of more efficient models but also it is generating more accurate neural networks compared to others conceived from the beginning. The high performance of this --longer path-- approach may be supported by the fact that effective transfer knowledge is being achieved by composing a final model from the most meaningful components of a larger and well-performing initial model.

In summary, based on the empirical results, we believe that optimizers derived from back propagation are not powerful enough to find an optimal function given a neural network topology (which represents a functional subspace). In consequence, methods optimizing dense feed-forward neural networks, like the one we have proposed, might be the best way to lead the data scientist to design efficient and effective deep neural networks for a learning task.

\section{Conclusion}
Deep learning models have been widely used during the last decade due to their learning and generalization capacities. However, these neural networks entail significant energetic costs and are hard to design efficiently. In this paper, we propose ODF2NNA, a new building method for feed-forward neural networks based on a three-step process: constructing a general model to address the learning task, train it properly and refine the model by finding the useful units per layer, removing the irrelevant ones (according to a parameter $\epsilon$) and finally, lightly retraining the new model. The proposal has been evaluated through a thorough empirical study on both classification and regression problems, including small, medium and large-scale datasets and learning tasks and comparing with 15 different techniques for pruning feed-forward neural networks. The experimental results confirm that simpler and more accurate models can be obtained. So the proposed method does not only allow to define effective topologies but drives to build well-performing neural networks based on knowledge transfer from relevant subnetworks. 

\section*{Acknowledgment}

This research has been partially supported by the Spanish Ministry of Economy, Industry and Competitiveness (TIN2016-81113-R and PID2020-118224RB-I00) and the Industry Andalusian Council (Consejería de Transformación Económica, Industria, Conocimiento y Universidades de la Junta de Andalucía, P18-TP-5168), with the cofinance of the European Union (FEDER).

\end{document}